\documentclass[10pt,twocolumn,letterpaper]{article}

\usepackage[pagenumbers]{cvpr} 

\usepackage{graphicx}
\usepackage{amsmath}
\usepackage{amssymb}
\usepackage{subcaption}
\usepackage{enumerate}
\usepackage{ulem}
\usepackage{makecell}
\usepackage{times,graphicx,amsmath,amssymb,booktabs,tabulary,multirow,overpic,xcolor}
\usepackage[pagebackref=true,breaklinks=true,letterpaper=true,colorlinks,bookmarks=false]{hyperref}

\usepackage[british,english,american]{babel}

\newcommand{\app}{\raise.17ex\hbox{$\scriptstyle\sim$}}

\newcolumntype{x}[1]{>{\centering\arraybackslash}p{#1pt}}

\newlength\savewidth\newcommand\shline{\noalign{\global\savewidth\arrayrulewidth
\global\arrayrulewidth 1pt}\hline\noalign{\global\arrayrulewidth\savewidth}}
\newcommand{\tablestyle}[2]{\setlength{\tabcolsep}{#1}\renewcommand{\arraystretch}{#2}\centering\footnotesize}
\makeatletter\renewcommand\paragraph{\@startsection{paragraph}{4}{\z@}
{.5em \@plus1ex \@minus.2ex}{-.5em}{\normalfont\normalsize\bfseries}}\makeatother

\usepackage[capitalize]{cleveref}
\crefname{section}{Sec.}{Secs.}
\Crefname{section}{Section}{Sections}
\Crefname{table}{Table}{Tables}
\crefname{table}{Tab.}{Tabs.}

\def\cvprPaperID{***} 
\def\confName{CVPR}
\def\confYear{2022}

\begin{document}

\title{SEA: Bridging the Gap Between One- and Two-stage Detector Distillation via SEmantic-aware Alignment}
\author{	Yixin Chen$^{1}$, Zhuotao Tian$^{1}$, Pengguang Chen$^{1}$, Shu Liu$^{2}$, Jiaya Jia$^{1,2}$  \\[0.2cm]
	The Chinese University of Hong Kong$^{1}$\quad SmartMore$^{2}$
}
\maketitle

\begin{abstract}
	We revisit the one- and two-stage detector distillation tasks and present a simple and efficient semantic-aware framework to fill the gap between them. We address the pixel-level imbalance problem by designing the category anchor to produce a representative pattern for each category and regularize the topological distance between pixels and category anchors to further tighten their semantic bonds. We name our method SEA (SEmantic-aware Alignment) distillation given the nature of abstracting dense fine-grained information by semantic reliance to well facilitate distillation efficacy.
	SEA is well adapted to either detection pipeline
	and achieves new state-of-the-art results on the challenging COCO object detection task on both one- and two-stage detectors. Its superior performance on instance segmentation further manifests the generalization ability. Both 2x-distilled RetinaNet and FCOS with ResNet50-FPN outperform their corresponding 3x ResNet101-FPN teacher, arriving 40.64 and 43.06 AP, respectively. Code will be made publicly available.  
\end{abstract}

\section{Introduction}
\label{sec:intro}

In object detection, well-designed backbones \cite{resnet, mobilev2, swin, vgg, efficientnet} provide strong support for powerful detectors to tackle challenging tasks \cite{coco, voc}. The balance between latency and accuracy is an inevitable trade-off in detection, especially when deploying models on mobile devices. Unlike pruning \cite{prunesonghan, deepcompress} and quantization \cite{quantjacob, dorefa, xnor, deepcompress, fullyquantdet} that alter the original model weights and storage bits, knowledge distillation \cite{Hintondistill, akd, dsig, chenkd, ckd, mutual, fgfi, attentiontransfer, tkd, defeat, gid, skd} instead keeps the target model intact by transferring knowledge from a well-trained teacher model to relatively small student model, which now becomes a common practice in model acceleration.
 
\begin{figure}[t]
	\begin{center}
		\includegraphics[width=1.0\linewidth, height=0.4\linewidth]{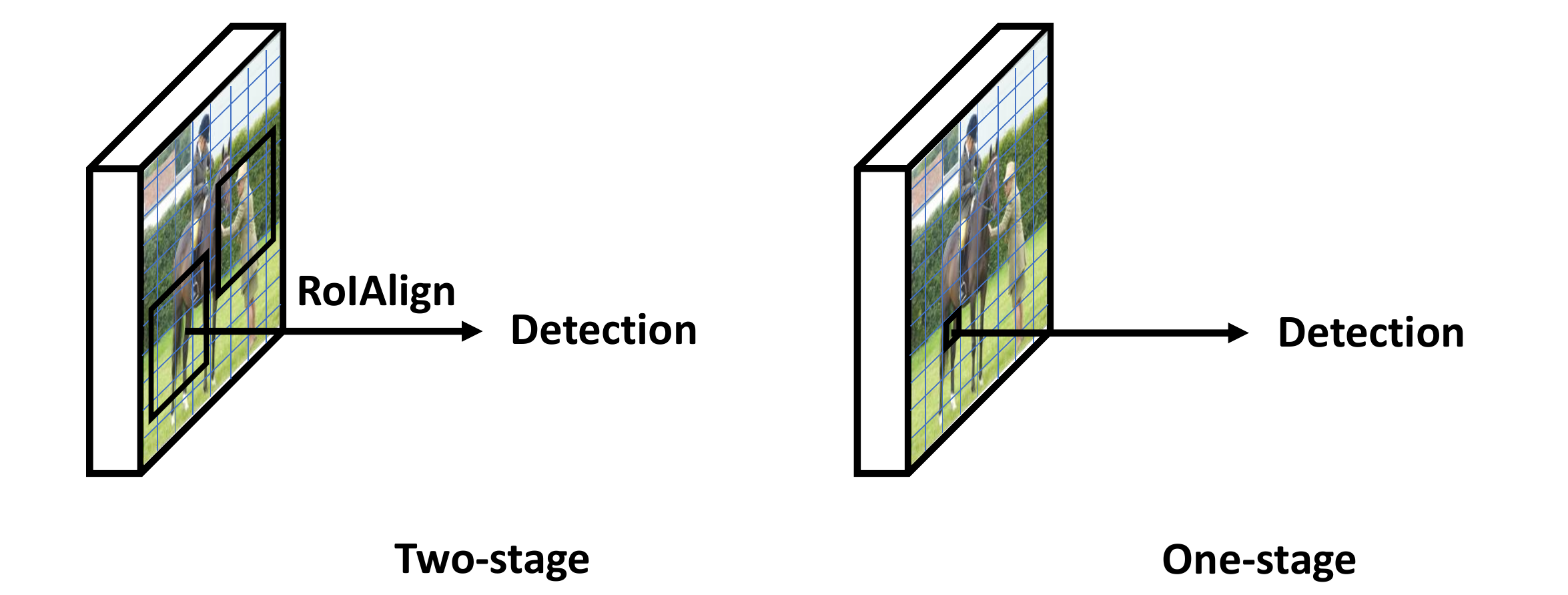}
	\end{center}
	\vspace{-0.25in}
	\caption{Difference between one- and two-stage detection.}
	\label{fig:one_two_stage}
	\vspace{-0.2in}
\end{figure}

Recently, one-stage detectors are favorable to mobile applications 
owing to free of specialized operators which are hard to eliminate. 
However, the distillation performance of one-stage detectors still fall behind with the two-stage ones. 
The reason is that most of seminal distillation methods \cite{dsig, chenkd, fgfi, tkd, defeat, gid} for object detection were mainly designed for two-stages detectors. 
For example, DSIG \cite{dsig} improves two-stage detectors \cite{fasterrcnn} by more than 2\%. But it decreases performance by 0.8\% when applied to FCOS \cite{fcos}.

This inconsistency mainly comes from the discrepancy between one- and two-stage detection pipelines (Figure~\ref{fig:one_two_stage}). In two-stage pipelines, prior proposals are generated by Region Proposal Network \cite{fasterrcnn} to extract region of interest (RoI) features, making prediction stem from pixel collections. While in one-stage counterparts (also known as dense detectors) without proposals, dense feature maps are processed by consecutive convolutions. Individual pixels are directly used to yield predictions. 

A potential issue existing in one-stage detector is that the ``interested regions" (\ie, selected individual pixels) are not aligned with the real interested regions (\ie, bounding boxes) in dense detectors. In contrast, the region proposals adopted in the two-stage designs are naturally suitable for instance-level box predictions. 
Besides, predictions of unimportant or wrong pixels might introduce unnecessary noise during the pixel-to-pixel distillation process between one-stage teacher and student models, 
such that the relation-based methods~\cite{gid, dsig} could erroneously leverage the correlation between elements, leading to performance deduction. 

To address this issue, we propose a new perspective to understand knowledge distillation for dense detectors. As mentioned before, dense detectors are comprised of multiple levels of feature maps, where pixel-level training samples are extremely imbalanced and no explicit instance-level relations can be exploited from the dense feature maps. 
Furthermore, dense detectors have another parallel branch to deal with localization information, which can be much better exploited. 

\paragraph{Resolve Drastic Imbalance:}
Different from two-stage detectors that regularize the ratio of foreground (FG) samples to background (BG) samples, one-stage ones just let it be. This results in a higher level of imbalance between FGs and BGs, and even between different class of foregrounds. Two-stage oriented distillers \cite{dsig, defeat} tend to decouple FGs from BGs and adjust their loss weights differently. In one-stage detectors, FGs and BGs are hard to be split and tuned on dense feature maps. The drastic imbalance makes distillation less effective if regularization is exerted. To address this problem, we encode a new knowledge representation called \textit{Category Anchor}, which is designed as a general pattern for each existing category. It also acts as semantic summary for dense feature maps of scene images. When it comes to distilling the \textit{Category Anchor}, the imbalance between categories (including BG) is moderated.  

\paragraph{Excavate Sparse Relations in Pixel-Level:}
Two-stage style instance-level relations \cite{dsig, gid} are difficult to be transferred to one-stage pipelines. The uniformly-sampled dense relation distillation \cite{skd} even worsens the detection performance. This indicates that the relations need to be better defined regarding the rich information inside each semantic pixel. Also, it will be better if irrelevant dense relations can be screened. 

To this end, we map the spatial pixel representation into a unit hypersphere, where the spatial local similarity with the nearby elements is retained without losing the discrimination ability with those having distinct characteristics. Instead of measuring the relations among all of them, we only quantify the distances between these elements and \textit{Category Anchors}. It lets plenty of meaningless relations be discarded and sparsification automatically complete (Figure~\ref{fig:anchor_distance}). We name this strategy \textit{Topological Distance Distillation}.     

\paragraph{Different Semantics in Disentangled Branches:}
The classification and bounding box regression are empirically not complementary (Figure~\ref{fig:clsbbox}). Thus they are accomplished independently by two separated branches, in order to disentangle features and make them focus on optimizing their own tasks. Moreover, experimental results (Table \ref{table:ablation}) show that simply matching pixels via MSE loss in bounding box branch is not conducive to the performance. 

As demonstrated in Figure~\ref{fig:clsbbox}, the high activation is generally clustered around objects in the branch for regressing the bounding box, while the activations of the classification branch are more likely to get widely spread starting from the object regions. Considering this discrepancy, instead of adopting simple pixel-pixel matching, we develop \textit{Localization Distribution Alignment} to match relative localization regression between student and teacher via probability distributions, to be detailed in Sec~\ref{sec:locloss}. In this way, we avoid distilling the absolute value of each activation map and find another approach to model information of relative locations in spatial domain.

\paragraph{Our Contributions:} We address the aforementioned issues and come up with the final solution for rarely studied one-stage detector distillation. To leverage the dense feature maps in multiple scales and layers, we design a proper knowledge representation \textit{Category Anchor} to summarize the general semantic pattern in the scene. Based on it, \textit{Topological Distance} is introduced to maintain the semantic bonds within pixel-level samples. Lastly, we form the localization distillation as a distribution alignment problem to efficiently transfer knowledge of localization information. We name our method Semantic-aware Alignment distillation regarding the nature to exploit fine-grained knowledge from a semantic-aware perspective.

Our method outperforms all previous state-of-the-art object detection distillers on the COCO \cite{coco} detection task by a large margin \wrt diverse modern one-stage detectors with heterogeneous backbones and shows its generality on the COCO instance segmentation task. In addition, we achieve strong performance on par with state-of-the-arts on two-stage detectors. While two-stage and one-stage detector adopt different processing scheme, they both share one common ground: the majority of parameters are devoted to yielding high-level semantic feature maps, which serve as the foundation for subsequent tasks. Distilling on this intersection is the key to bridging the gap between them.  
In our ablation study, we analyze the effect of each component and test hyper-parameters sensitivity. Our method is stable to train and does not introduce extra parameters to distillation training. To the best of our knowledge, this is the first attempt of realizing one-stage detector distillation.

\section{Related Work}
\label{sec:related}

\begin{figure*}[!t]
	\begin{center}
		\includegraphics[height=0.45\linewidth,width=1.0\linewidth]{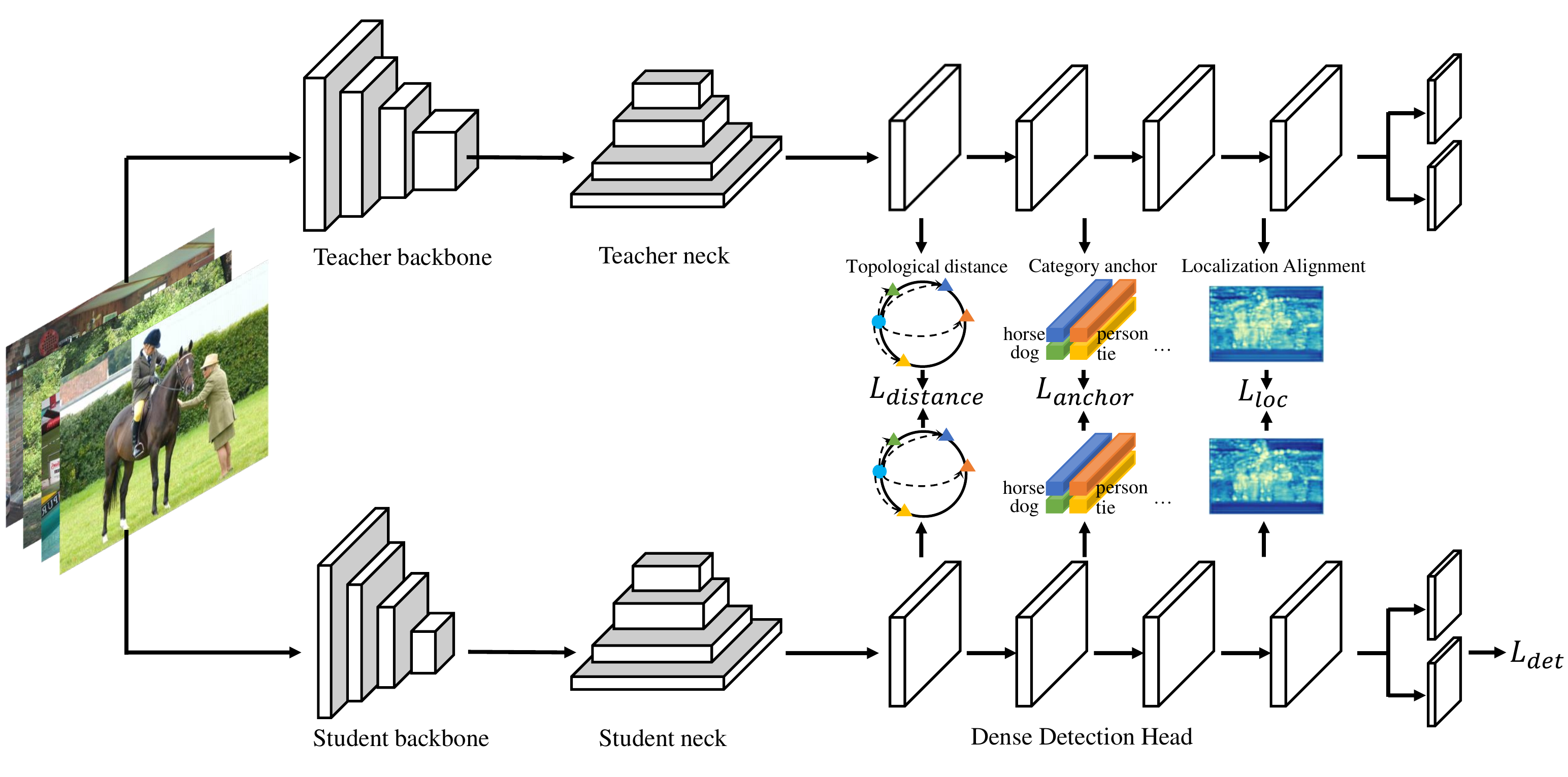}
	\end{center}
	\vspace{-0.32in}
	\caption{Distillation framework of our method. For simplicity, we only display one branch between the classification and bounding box branches in the dense detection head. $L_{anchor}$ and $L_{distance}$ align the category anchors and topological distances generated in each feature map from student and teacher, respectively, and $L_{loc}$ aligns the feature map in the bounding box branch.}
	\label{fig:diagram}
	\vspace{-0.2in}
\end{figure*}

\subsection{Object Detectors}

Deep learning based object detectors are categorized into two groups of two-stage and one-stage detectors. The former generates prior proposals to extract regional features, while the latter accomplishes detection without the priors.

Two-stage methods are mainly originated from R-CNN \cite{rcnn}, the following works \cite{fastrcnn, fasterrcnn, cascade} further refined the R-CNN pipeline to generate more precise proposals and save the implemented real-time.  
There are also many excellent methods \cite{doublehead, htc, dynamicrcnn} to further improve two-stage detectors.
They all enclose the extracted features by regions and treat them as instances in classification and localization.

In the absence of generated proposals, one-stage detectors \cite{yolo, yolof, retina, atss, gfl, fcos, ssd} were designed for low latency while maintaining high accuracy. YOLO \cite{yolo} was first proposed to introduce the anchor boxes to predict classes and bounding boxes from feature maps. In order to alleviate the foreground-background imbalance issue in dense detection, RetinaNet \cite{retina} proposed the focal loss to down-weight the well-classified easy examples. Further, ATSS \cite{atss} refined the label assign algorithm and GFL \cite{gfl} proposed an optimized focal loss and a box distribution loss. There are also anchor-free detectors that discard the bounding box prior, such as FCOS \cite{fcos}, which implemented detection from only points. In our paper, we experiment with these detectors, including two-stage \cite{fasterrcnn, cascade}, one-stage anchor-based \cite{retina, atss, gfl} and anchor-free \cite{fcos} ones.         

\subsection{Knowledge Distillation}
Knowledge Distillation transfers knowledge from a large capacity teacher model to a light weight student model. Hinton \etal \cite{Hintondistill} first regarded the soft logits of output layers as knowledge representation to be distilled. Besides, intermediate feature representations \cite{fitnet} and attention maps \cite{attentiontransfer} are also used to match student and teacher. Recently, several knowledge distillation frameworks further excavate the potential, such as Mutual Learning \cite{mutual}, Noisy Student \cite{noisystu}, and TAKD \cite{kdta}.      

Object Detection benefits a lot from knowledge distillation since it achieves considerable accuracy with less inference cost and memory footprint. We categorize detection distillation methods into four groups: 

(1) Region-based: 
Chen \etal \cite{chenkd} first presented a detection knowledge distillation framework for two-stage detectors and specially re-designed the logits distillation loss for detection.  
Wang \etal \cite{fgfi} generated the fine-grained masks from ground truth to distill the selected near-object regions in feature maps. 
Sun \etal \cite{tkd} proposed a Gaussian mask to achieve more attention on regions near object center and learning rate decay strategy to improve generalization. Recently, Guo \etal \cite{defeat} split the features into foreground and background, treating them differently.
These methods all first extract attentional regions and then conduct distillation on these selected pixels.

(2) Relation-based:
Chen \etal \cite{dsig} and Dai \etal \cite{gid} utilized the instance relations in detection to align student with teacher. These methods also use prior regions (\ie proposals and ground-truth boxes) to select regional features as elements to build relations for further distillation.  

(3) Backbone-based:
Zhang \etal \cite{akd} leveraged the attention module and generated a level of attention maps for conducting backbone distillation in detector. However, it does not utilize the following detection head, which also contains sufficient information.

(4) Segmentation-based:
There also exist methods derived from the segmentation task. Liu \etal \cite{skd} proposed a structured method to transfer relations between regular block of pixels. Note in detection, this kind of relation brings much noise due to high ratio of background. Shu \etal \cite{ckd} distilled channel-wise information -- the activation map of detectors is dissimilar from that of segmentation.

\section{Semantic-aware Alignment Distillation}
\label{sec:method}

In this section, we introduce our overall distillation pipeline that is composed of \textit{category anchor distillation}, \textit{topological distance distillation} and \textit{localization distribution alignment}. The category anchors represent the global categorical information mined from multi-scale feature maps. 
It depicts the distribution between semantic prototypes and individual pixels without being adversely affected by the inter-class imbalance. Thus the topological distance can be better illustrated in latent embedding space.
The localization distribution alignment effectively utilizes distribution matching loss to distill the regression feature maps from teacher to student, where pixels exhibit semantics in spatially relative locations. 

Our diagram is shown in Figure~\ref{fig:diagram}, we apply a set of distillation losses on the dense detector head (convolutional branches), which does not introduce additional parameter and brings minor extra training overhead. In our pipeline, anchors for individual categories are firstly generated from feature maps. Based on that, the topological distance between individual pixel and anchor is computed. 
With student and teacher anchors and the topological distances, the anchor loss $L_{anchor}$ and topological distance loss $L_{distance}$ are applied to student training. The location distribution alignment loss $L_{loc}$ regularizes the bounding box branch. We note that no distillation is applied yet to the detector backbone, leaving much room for further improvement in our future work.

\subsection{Category Anchor Distillation}
\begin{figure}[!t]
	\begin{center}
		\includegraphics[width=1.0\linewidth]{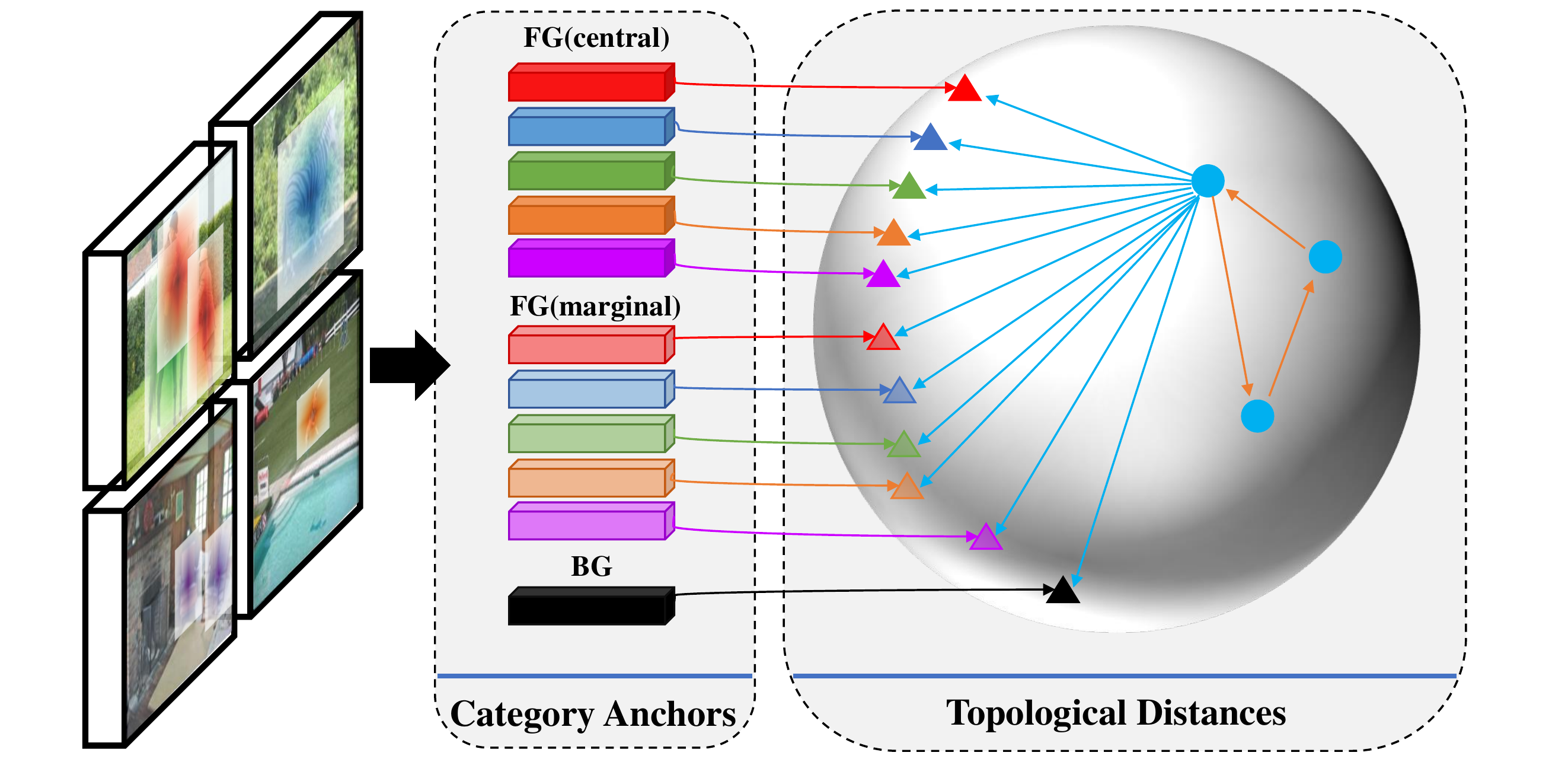}
	\end{center}
	\vspace{-0.2in}
	\caption{Category anchors and Topological Distances. We trim the outermost circle of pixels in the instance region and define them to be the marginal part, while the remainder in this region is the central part. We map anchors ($\triangle$) and pixels (\textcolor{cyan}{$\bullet$}) into a unit hypersphere, on which the distances are computed. We only distill distances between pixel and anchors (\textcolor{cyan}{$\rightarrow$}) and ignore the distances among points (\textcolor{orange}{$\rightarrow$}) to achieve the sparsification. }
	\label{fig:anchor_distance}
	\vspace{-0.10in}
\end{figure}

To avoid imbalanced information from a series of convolutional layers by dense pixel matching, we devise category anchor as categorical summary of existing instances in one image batch (Figure~\ref{fig:anchor_distance}). A categorical summary functions like the two-stage detectors that extract regions of interest to assist detection. Instead of pixel-wise imitation between the student and teacher models like \cite{defeat, fgfi}, we elect attentional pixels to build category anchors and make dense feature maps focus on categorical regions. Considering that square bounding boxes do not represent actual shape of real objects, we divide boxes into the central and marginal parts. Thus, each category owns two category anchors, which collect all pixels belonging to the categorical regions across the whole image batch.   

Specifically, in one image batch $\mathcal{I}$ with feature map $\mathcal{F}$ produced by layer $\mathcal{L}$, we denote category anchors as $C=\{c_{1},..., c_{2i-1}, c_{2i}, c_{bg}\}$, where $i$ indexes different categories except the background in this batch and 2$i$-1, 2$i$ here denotes the central, marginal category anchors for category $i$, respectively. Note that, we do not discard pixels belonging to the background like \cite{dsig, fgfi} and instead produce an anchor for the background. 
The category anchors are generated through the following steps. 
\begin{enumerate}
	\vspace{-1mm}
	\item Initialize all-zeros Category masks $\mathcal{M}=\{M_{1},...,M_{2i},M_{bg} \}$. For each instance in $\mathcal{I}$ belonging to category $j$ ($1 \!\leq\! j \!\leq\! i$), we split it into central and marginal part and set the values of corresponding regions on $M_{2j-1}$, $M_{2j}$ respectively to 1. The final Category masks turn out to be $\mathcal{M} \!=\! \{M_{1},...,M_{2i},M_{bg}\}$, where $M_{bg} \!=\! \overline{\sum_{j}^{i} M_{2j-1}\!+\! M_{2j}}$.
	\vspace{-0.5mm}
	\item For the $p$-th ($1 \!\leq\! p \! \leq\! 2i+1$) category anchor, the relevant regions on the feature maps are masked by $M_{p}$: $\mathcal{F}_{p}$ = $\mathcal{F} * M_{p}$.
	\vspace{-0.5mm}
	\item For the $p$-th category anchor, the masked feature maps are vectorized to category anchors by spatial average pooling as $c_{p}$ =  AvgPool2d($\mathcal{F}_{p}$).
	\vspace{-1mm}
\end{enumerate}      
With the multi-level detection scheme proposed in \cite{fpn}, networks with consecutive convolutional layers also yield multi-scale feature maps. Therefore, in our distillation framework, the final category anchors are denoted as $\mathcal{C}=\{C_{1,1},...,C_{m,k}\}$, where $C_{m,k}$ is the category anchor generated from the $m$-th conv with $k$-th scale-level of the convolutional feature maps. 
More details are in supplementary.

\paragraph{Multi-level Anchor Matching:}


Different levels of anchors concentrate information collected from multi-resolution feature maps. We distill student and teacher anchors on all levels of feature maps as 
\begin{equation}
	L_{anchor} = \frac{1}{N_{m}N_{k}} \sum_{m}^{N_m} \sum_{k}^{N_k} 1 -
	\frac{C_{m, k}^s\cdot C_{m, k}^t}{\|C_{m, k}^s\|\cdot \|C_{m, k}^t\|},
\end{equation}
where $C^{s}$ and $C^{t}$ represent the anchors of student and teacher, respectively. $L_{anchor}$ aims to align each level of student category anchors with that of teacher. It makes sure optimization towards the general category prototype from a multi-scale perspective. We apply it to both classification and bounding box branches.

\subsection{Topological Distance Distillation}

We regard each pixel as an individual sample in one-stage detection framework where each pixel owns adequate information to support the following classification and localization regression tasks. 
Similarly, samples scatter in embedding spaces and have their correlations with each other \cite{skd}. Rather than modeling dense correlations among all pixels, we measure the distances between pixels and category anchors. The distances construct a topological structure for calibrating individual samples against the anchors. Well trained teacher exhibits more precise measure regarding the distances. Thus it regularizes student topology. We devise the topological distance distillation loss between student and teacher as
\begin{equation}\label{eq:distanceloss}
	\begin{split}
		L_{distance}\! =\! \frac{1}{N_{m}N_{k}} \sum_{m}^{N_m} \sum_{k}^{N_k} D(\mathcal{F}_{m,k}^{s} \!\circledast\! C_{m,k}^{s} \| \mathcal{F}_{m,k}^{t} \!\circledast\! C_{m,k}^{t}) \\
		\!\!=\!\frac{1}{N_{m}N_{k}} \sum_{m}^{N_j} \sum_{k}^{N_k} p(\frac{\mathcal{F}_{m,k}^{s} \!\circledast\! C_{m,k}^{s}}{\tau_d}) \log \frac{p(\frac{\mathcal{F}_{m,k}^{s} \circledast C_{m,k}^{s}}{\tau_d})}{q(\frac{\mathcal{F}_{m,k}^{t} \circledast C_{m,k}^{t}}{\tau_d})} 
	\end{split}
\end{equation}           
where $\mathcal{F} \circledast C $ denotes the cosine similarity for computing the distances and $\tau_d$ stands for the temperature.  

\subsection{Localization Distribution Alignment}
\label{sec:locloss}

\begin{figure}[t]
	\begin{center}
	\begin{subfigure}{0.45\linewidth}
		\centering
		\includegraphics[width=\linewidth]
		{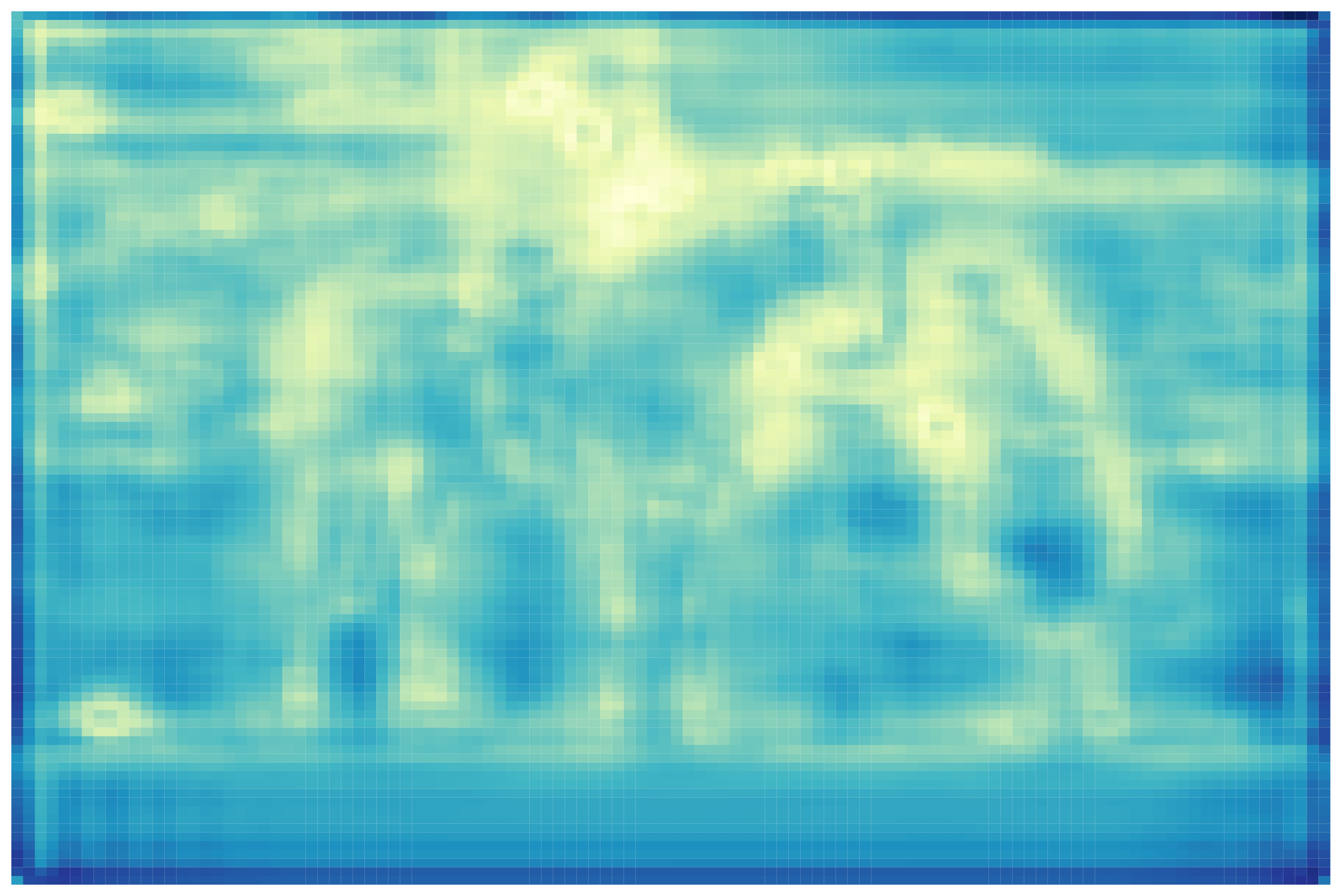}
		\caption{Classification branch}
	\end{subfigure}
	\begin{subfigure}{0.45\linewidth}
		\centering
		\includegraphics[width=\linewidth]
		{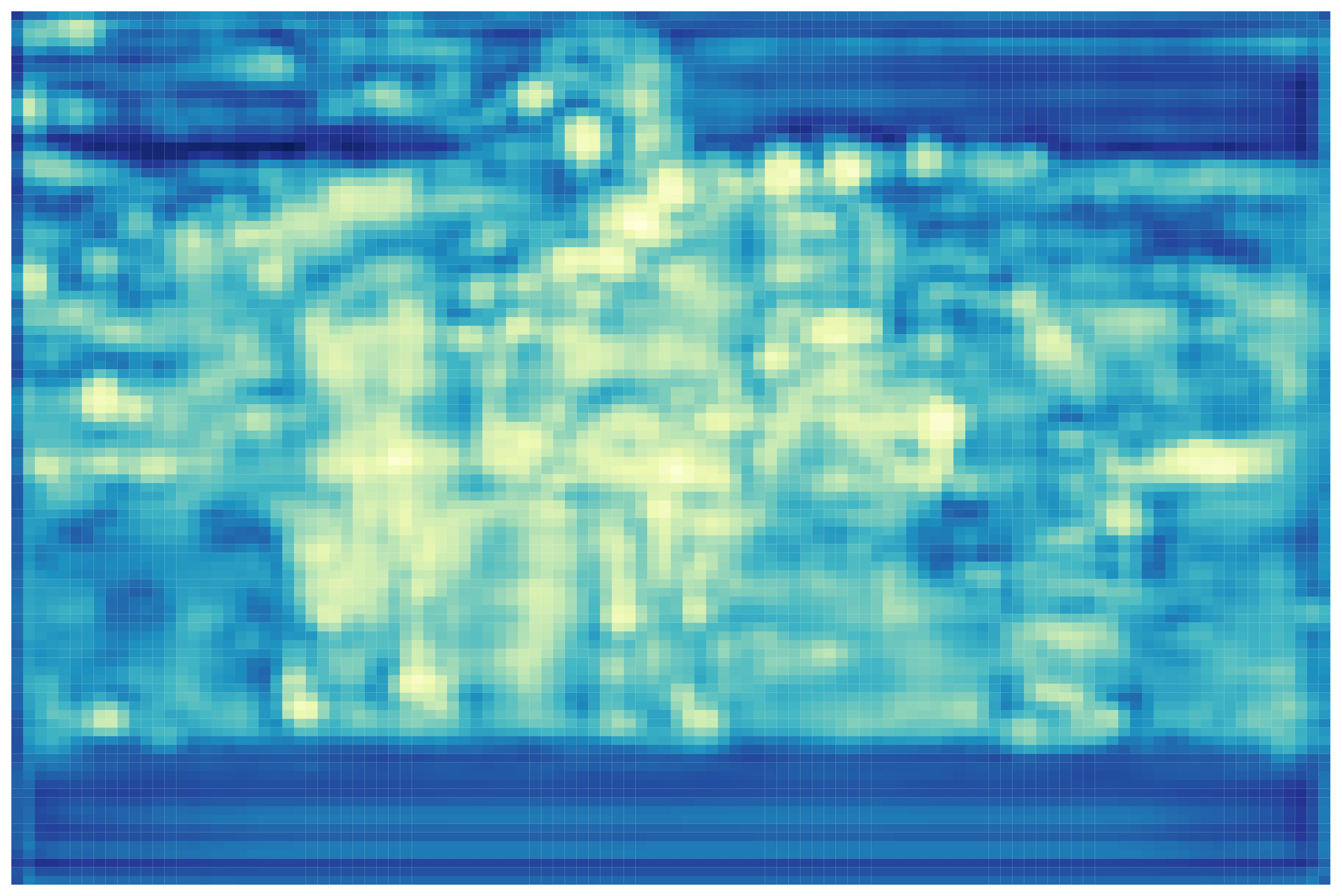}
		\caption{Bounding box branch}
	\end{subfigure}
	\end{center}
	\vspace{-0.2in}
	\caption{Difference in the activation map between classification and bounding box branches. We use the feature map produced by the last convolution of RetinaNet-R101.}
	\vspace{-0.2in}
	\label{fig:clsbbox}
\end{figure}

\begin{table*}[t]
	\tablestyle{3.5pt}{1.05}
	\begin{minipage}[!t]{\columnwidth}
		\begin{tabular}{l|x{15}x{15}|x{28}|x{14}x{14}x{18}|x{14}x{14}x{14}}
			& Stu & Tch & Schedule & AP & AP$_{50}$ & AP$_{75}$ & AP$_S$ & AP$_M$ & AP$_L$\\
			\shline
			RetinaNet  & R18 & - & 1x
			& 31.60 & 49.61 & 33.36 & 17.06 & 34.80 & 41.11 \\
			RetinaNet  & R18 & R50 & 1x
			& \textbf{35.94} & 54.12 & 38.34 & 19.36 & 39.28 & 47.79\\
			RetinaNet  & MV2 & - & 1x
			& 28.68 & 46.69 & 30.05 & 16.01 & 31.12 & 37.40\\
			RetinaNet  & MV2 & R50 & 1x
			& \textbf{33.99} & 51.67 & 36.18 & 18.14 & 37.58 & 45.96 \\
			RetinaNet  & - & R50 & 3x
			& 38.67 & 57.99 & 41.48 & 23.34 & 42.30 & 50.31 \\
			\hline
			RetinaNet  & R50 & - & 1x
			& 37.22 & 56.58 & 39.90 & 22.48 & 41.60 & 47.40 \\
			RetinaNet  & R50 & R101 & 1x
			& \textbf{39.91} & 59.80 & 42.91 & 22.88 & 44.04 & 52.34 \\
			RetinaNet  & EB0 & - & 1x
			& 33.35 & 53.53 & 35.21 & 19.07 & 36.30 & 44.43 \\
			RetinaNet  & EB0 & R101 & 1x
			& \textbf{36.61} & 55.94 & 38.81 & 19.39 & 40.28 & 50.10 \\
			RetinaNet  & - & R101 & 3x
			& 40.39 & 60.25 & 43.18 & 24.02 & 44.34 & 52.18 \\
			\hline
			\hline
			ATSS  & R18 & - & 1x
			& 34.81 & 51.99 & 37.38 & 19.53 & 38.01 & 44.79 \\
			ATSS  & R18 & R50 & 1x
			& \textbf{38.24} & 55.99 & 41.22 & 21.30 & 41.53 & 51.08 \\
			ATSS  & MV2 & - & 1x
			& 31.20 & 47.44 & 33.36 & 16.81 & 33.78 & 41.08 \\
			ATSS  & MV2 & R50 & 1x
			& \textbf{34.85} & 51.88 & 37.31 & 19.21 & 37.76 & 46.27 \\
			ATSS  & - & R50 & 3x
			& 42.04 & 60.24 & 45.45 & 25.74 & 45.65 & 54.16 \\
			\hline
			ATSS  & R50 & - & 1x
			& 39.75 & 57.77 & 43.30 & 22.78 & 43.93 & 51.70 \\
			ATSS  & R50 & R101 & 1x
			& \textbf{42.17} & 60.64 & 45.71 & 25.65 & 46.10 & 55.02 \\
			ATSS  & EB0 & - & 1x
			& 35.52 & 53.21 & 38.16 & 20.16 & 38.28 & 46.38 \\
			ATSS  & EB0 & R101 & 1x
			& \textbf{38.40} & 56.65 & 41.26 & 21.63 & 41.84 & 51.65 \\
			ATSS  & - & R101 & 3x
			& 43.59 & 62.10 & 47.31 & 27.67 & 47.67 & 56.37 \\	
		\end{tabular}
	\end{minipage}
	\hspace{15pt}
	\begin{minipage}[!t]{\columnwidth}
		\begin{tabular}{l|x{15}x{15}|x{28}|x{14}x{14}x{18}|x{14}x{14}x{14}}
			& Stu & Tch & Schedule &  AP &  AP$_{50}$ & AP$_{75}$ & AP$_S$ &  AP$_M$ &  AP$_L$\\
			\shline
			FCOS  & R18 & - & 1x
			& 33.99 & 51.92 & 36.33 & 20.04 & 36.82 & 43.41 \\
			FCOS  & R18 & R50 & 1x
			& \textbf{38.14} & 56.56 & 40.78 & 22.23 & 41.19 & 49.16 \\
			FCOS  & MV2 & - & 1x
			& 30.73 & 47.64 & 32.77 & 17.75 & 32.92 & 39.90 \\
			FCOS  & MV2 & R50 & 1x
			& \textbf{34.94} & 52.70 & 37.53 & 20.41 & 37.87 & 45.47 \\
			FCOS  & - & R50 & 3x
			& 41.28 & 60.03 & 44.57 & 25.93 & 44.97 & 52.06 \\
			\hline
			FCOS  & R50 & - & 1x
			& 39.34 & 58.09 & 42.64 & 24.16 & 43.37 & 49.77 \\
			FCOS  & R50 & R101 & 1x
			& \textbf{42.52} & 61.04 & 45.96 & 25.90 & 46.61 & 55.02 \\
			FCOS  & EB0 & - & 1x
			& 34.57 & 52.79 & 37.04 & 20.90 & 36.89 & 44.51 \\
			FCOS  & EB0 & R101 & 1x
			& \textbf{38.11} & 56.73 & 40.85 & 21.73 & 41.56 & 50.60 \\
			FCOS  & - & R101 & 3x
			& 42.88 & 61.69 & 46.16 & 27.09 & 46.54 & 54.85 \\
			\hline
			\hline
			GFL  & R18 & - & 1x
			& 35.94 & 53.20 & 38.51 & 19.92 & 38.93 & 47.45 \\
			GFL  & R18 & R50 & 1x
			& \textbf{38.64} & 56.45 & 41.56 & 21.11 & 42.20 & 51.42 \\
			GFL  & MV2 & - & 1x
			& 32.49 & 48.93 & 34.64 & 16.79 & 35.26 & 43.02 \\
			GFL  & MV2 & R50 & 1x
			& \textbf{36.46} & 53.92 & 39.01 & 19.57 & 39.65 & 49.24 \\
			GFL  & - & R50 & 3x
			& 42.38 & 60.83 & 45.36 & 25.23 & 45.82 & 55.43 \\
			\hline
			GFL  & R50 & - & 1x
			& 40.30 & 57.96 & 43.56 & 23.46 & 44.37 & 52.67 \\
			GFL  & R50 & R101 & 1x
			& \textbf{42.83} & 61.36 & 46.44 & 24.68 & 46.84 & 57.15 \\
			GFL  & EB0 & - & 1x
			& 37.53 & 55.15 & 40.55 & 20.61 & 40.44 & 51.31 \\
			GFL  & EB0 & R101 & 1x
			& \textbf{38.78} & 56.64 & 41.73 & 20.50 & 42.45 & 53.73 \\
			GFL  & - & R101 & 3x
			& 44.13 & 62.75 & 47.70 & 26.60 & 48.09 & 57.56 \\
		\end{tabular}
	\end{minipage}
	\vspace{-2mm}
	\caption{Our distilled \textbf{Object Detection} \emph{box} AP on COCO \texttt{val} \wrt one-stage detectors. We validate it on anchor-based RetinaNet \cite{retina}, ATSS \cite{atss}, GFL \cite{gfl} and anchor-free FCOS \cite{fcos} with various backbones of ResNet18(\textbf{R18}), ResNet50(\textbf{R50}), ResNet101(\textbf{R101}), MobileNetV2(\textbf{MV2}) and EfficientNet-B0(\textbf{EB0}).}
	\label{table:onestage}
	\vspace{-3mm}
\end{table*}

We empirically observe that the bounding box branch in detectors behaves differently from the classification branch. It is noticed that the pixels in the classification branch are more activated on the edge of ground truth instances, revealing highly salient semantics. However, for the bounding box (bbox) branch where the content of each pixel is not correlated with class information, it is less semantic-sensitive about scene understanding since it predicts the deltas from anchor box or point to the ground truth boxes. 

As shown in Figure~\ref{fig:clsbbox}, the bbox feature maps have pixels near an object center in the image more activated, which actually indicates the relative locations between itself and the ground truth center. Inspired by this finding, we model distribution alignment in spatial domain, and effectively adopt the Kullback-Leibler divergence loss to distill the bbox layer from teacher to student as
\begin{equation}\label{eq:locloss}
	\begin{split}
		L_{loc} &= \frac{1}{N_{m}N_{k}} \sum_{m}^{N_m} \sum_{k}^{N_k} D(\mathcal{F}_{m,k}^{s} \| \mathcal{F}_{m,k}^{t}) \\
		&= \frac{1}{N_{m}N_{k}} \sum_{m}^{N_m} \sum_{k}^{N_k} p(\frac{\mathcal{F}_{m,k}^{s} }{\tau_l}) \log \frac{p(\frac{\mathcal{F}_{m,k}^{s} }{\tau_l})}{q(\frac{\mathcal{F}_{m,k}^{t} }{\tau_l})}
	\end{split}
\end{equation}
where $\mathcal{F}_{m,k}$ denotes the $m$-th scale and $k$-th bbox branch feature map. $\tau_l$ is the logit temperature.

\subsection{Overall Distillation Loss}

Incorporating the three aforementioned distillation losses into detection loss in student detector training, we yield the oveall distillation loss as
\begin{equation}\label{eq:overallloss}
	L = L_{det} + \lambda_a L_{anchor} + \lambda_d L_{distance} + \lambda_l L_{loc}
\end{equation}
where $L_{det}$ is the original detection loss for student detector, including classification and localization loss, which is designed differently in each type of detector. $\lambda_a$, $\lambda_d$ and $\lambda_l$ represent the penalty coefficients for $L_{anchor}$, $L_{distance}$ and $L_{loc}$, respectively. We empirically set $\{\lambda_a, \lambda_d, \lambda_l, \tau_d, \tau_l\}$ to $\{10, 1000, 1, 0.1, 0.1\}$. Detailed implementations of three proposed losses are provided in supplementary material.

\section{Experiments}
\label{sec:experiments}

\paragraph{Experimental Benchmark:} We adopt the most popular and challenging object detection dataset COCO\footnote{COCO images abide by Flickr Terms of Use and annotations are licensed under a Creative Commons Attribution 4.0 License.}\cite{coco} to evaluate the effectiveness of our proposed method and make comparison. 
We evaluate results with the detection average precision~(AP) over IoU threshold. The COCO style AP metrics, including mAP@[0.5:0.95], AP$_{50}$, AP$_{75}$, AP$_{S}$, AP$_{M}$, and AP$_{L}$, are applied.

\paragraph{Network Architectures:} 
Our base structure is on Detectron2 \cite{d2}.
In detail, we adopt ResNet-FPN as the backbone architecture for teacher. Simultaneously, we choose the ResNet-series \cite{resnet} ResNet18/50, mobile-level MobileNetV2 \cite{mobilev2} and NAS-based EfficientNet-B0 \cite{efficientnet} as backbones for the student. 
To verify the generality of our method, we distill on one-stage anchor-based RetinaNet \cite{retina}, ATSS \cite{atss}, GFL \cite{gfl}, one-stage anchor-free FCOS \cite{fcos}, and also two-stage Faster R-CNN \cite{fasterrcnn} and Cascade R-CNN \cite{cascade} with these backbones. 

\paragraph{Training Details:}\label{sec:training_detail} 
For existing detectors, we adopt the off-the-shelf pre-trained teacher from Detectron2 Model Zoo, and prepare other teacher detectors following the same training details as student. 
With the guidance of teacher models, we compose different architecture and capacity student-teacher pairs (R18-R50, R50-R101, MV2-R50, and EB0-R101) on aforementioned detectors. 
Detailed training setting is given in the supplementary material.

\subsection{Main Results}

We show our notable results on state-of-the-art one-stage detectors of RetinaNet \cite{retina}, FCOS \cite{fcos}, ATSS \cite{atss} and GFL \cite{gfl} with multiple heterogeneous backbones, as shown in Table~\ref{table:onestage}. 
Our method yields substantial improvement on average with four backbones: RetinaNet (+3.9 AP), FCOS (+3.8 AP), ATSS (+3.1 AP) and GFL (+2.6 AP).
Our method also brings huge promotion on small-capacity models. Distilled by ResNet-50, ResNet-18 student detectors gain 3.66 AP on average. Though guided by dissimilar architecture of the teacher network, MobileNetV2 student detectors still have 4.29 AP on average beyond the baseline. For larger detectors, ResNet-50 student detectors surpass the baseline by 2.71 AP on average. We also implement NAS-Architecture backbone EfficientNet-B0 on those detectors, which is improved by 2.73 AP on average.

We also validate our method on two-stage detectors of Faster R-CNN \cite{fasterrcnn} and Cascade R-CNN \cite{cascade}, as listed in Table~\ref{table:twostage}. With four distillation pairs, Faster R-CNN and Cascade R-CNN achieve 3.29 and 2.22 AP gain on average against student baseline, indicating that our method is surprisingly general. 

\begin{table}[t]
	\tablestyle{3.5pt}{1.0}
	\begin{tabular}{l|cc|c|x{22}}
		& Student & Teacher & Schedule &  AP$_{box}$ \\
		\shline
		Faster R-CNN  & R18 & - & 1x
		& 33.06 \\
		Faster R-CNN  & R18 & R50 & 1x
		& \textbf{36.95} \\
		Faster R-CNN  & MV2 & - & 1x
		& 29.47 \\
		Faster R-CNN  & MV2 & R50 & 1x
		& \textbf{34.15} \\
		Faster R-CNN  & R50 & - & 1x
		& 38.03 \\
		Faster R-CNN  & R50 & R101 & 1x
		& \textbf{40.63} \\
		Faster R-CNN  & EB0 & - & 1x
		& 33.85 \\
		Faster R-CNN  & EB0 & R101 & 1x
		& \textbf{35.82} \\
		\hline
		\hline
		Cascade R-CNN & R18 & - & 1x
		& 37.03 \\
		Cascade R-CNN & R18 & R50 & 1x
		& \textbf{40.70} \\
		Cascade R-CNN  & MV2 & - & 1x
		& 35.22 \\
		Cascade R-CNN  & MV2 & R50 & 1x
		& \textbf{37.52} \\
		Cascade R-CNN  & R50 & - & 1x
		& 41.57 \\
		Cascade R-CNN  & R50 & R101 & 1x
		& \textbf{43.79} \\
		Cascade R-CNN  & EB0 & - & 1x
		& 38.41 \\ 
		Cascade R-CNN  & EB0 & R101 & 1x
		& \textbf{39.11} \\ 
		
	\end{tabular}
	\vspace{-2mm}
	\caption{Our distilled \textbf{Object Detection} \emph{box} AP on COCO \texttt{val} \wrt two-stage Faster R-CNN \cite{fasterrcnn} and Cascade R-CNN \cite{cascade}.}
	\label{table:twostage}
	\vspace{-3mm}
\end{table}	

\begin{table}[t]
	\tablestyle{3.5pt}{1.0}
	\begin{tabular}{l|cc|c|x{22}|x{25}}
		& Student & Teacher & Schedule &  AP$_{box}$ & AP$_{mask}$ \\
		\shline
		Mask R-CNN  & R18 & - & 1x
		& 33.89 & 31.30 \\
		Mask R-CNN  & R18 & R50 & 1x
		& 37.17 & \textbf{33.94} \\
		Mask R-CNN  & MV2 & - & 1x
		& 30.65 & 28.55 \\
		Mask R-CNN  & MV2 & R50 & 1x
		& 34.54 & \textbf{31.62} \\
		Mask R-CNN  & R50 & - & 1x
		& 38.64 & 35.24 \\
		Mask R-CNN  & R50 & R101 & 1x
		& 40.82 & \textbf{36.99} \\
		Mask R-CNN  & EB0 & - & 1x
		& 35.28 & 32.81 \\
		Mask R-CNN  & EB0 & R101 & 1x
		& 37.93 & \textbf{34.53} \\
		\hline
		\hline
		SOLOv2  & R18 & - & 1x
		& - & 31.54 \\
		SOLOv2  & R18 & R50 & 1x
		& - & \textbf{33.71} \\
		SOLOv2  & MV2 & - & 1x
		& - & 28.45 \\
		SOLOv2  & MV2 & R50 & 1x
		& - & \textbf{31.26} \\
		SOLOv2  & R50 & - & 1x
		& - & 34.94 \\
		SOLOv2  & R50 & R101 & 1x
		& - & \textbf{37.08} \\
		SOLOv2  & EB0 & - & 1x
		& - & 31.42 \\
		SOLOv2  & EB0 & R101 & 1x
		& - & \textbf{33.63} \\
	\end{tabular}
	\vspace{-2mm}
	\caption{Our distilled \textbf{Instance Seg.} \emph{mask} AP on COCO \texttt{val} \wrt two-stage Mask R-CNN \cite{maskrcnn} and one-stage SOLOv2 \cite{solov2}.}
	\label{table:instseg}
	\vspace{-3mm}
\end{table}	

\subsection{Comparison with other Methods}

\begin{table*}[t]
	\tablestyle{3.5pt}{1.0}
	\begin{tabular}{l|l|cc|c|x{22}x{22}x{22}|x{22}x{22}x{22}}
		& Method & Student & Teacher & Schedule &  AP &  AP$_{50}$ & AP$_{75}$ & AP$_S$ &  AP$_M$ &  AP$_L$\\
		\shline
		RetinaNet & Baseline & R50 & - & 1x
		& 37.22 & 56.58 & 39.90 & 22.48 & 41.60 & 47.40 \\
		RetinaNet & FGFI \cite{fgfi} & R50 & R101 & 1x
		& 37.53 & 56.89 & 39.78 & 22.33 & 41.68 & 48.26 \\
		RetinaNet & TAD \cite{tkd} & R50 & R101 & 1x
		& 37.06 & 56.15 & 39.50 & 22.28 & 41.12 & 47.45 \\
		RetinaNet & CKD \cite{ckd} & R50 & R101 & 1x
		& 37.57 & 56.82 & 40.21 & 22.10 & 41.59 & 48.40 \\
		RetinaNet & SKD \cite{skd} & R50 & R101 & 1x
		& 37.62 & 57.02 & 40.15 & 22.60 & 41.82 & 48.20 \\
		RetinaNet & Defeat \cite{defeat} & R50 & R101 & 1x
		& 39.10 & 58.10 & 42.06 & 22.86 & 43.31 & 51.60 \\
		RetinaNet & DSIG \cite{dsig} & R50 & R101 & 1x
		& 39.25 & 59.07 & 42.04 & 22.96 & 43.57 & 51.44 \\
		RetinaNet & Ours & R50 & R101 & 1x
		& \textbf{39.91}  & 59.56 & 43.06 & 23.10 & 43.78 & 52.74  \\
		RetinaNet & $\dagger$GID \cite{gid} & R50 & R101 & 2x
		& 39.10 & 59.00 & 42.30 & 22.80 & 43.10 & 52.30 \\
		RetinaNet & AKD \cite{akd} & R50 & R101 & 2x
		& 39.60 & 58.80 & 42.10 & 22.70 & 43.3 & 52.5 \\
		RetinaNet & Ours & R50 & R101 & 2x
		& \textbf{40.52} & 60.55 & 43.59 & 24.57 & 44.42 & 52.53 \\
		\shline
		FCOS & Baseline & R50 & - & 1x
		& 39.34 & 58.09 & 42.64 & 24.16 & 43.37 & 49.77 \\
		FCOS & FGFI \cite{fgfi} & R50 & R101 & 1x
		& 39.38 & 57.96 & 42.54 & 23.77 & 43.50 & 49.97 \\
		FCOS & TAD \cite{tkd} & R50 & R101 & 1x
		& 39.15 & 57.83 & 42.25 & 23.89 & 42.96 & 50.00 \\
		FCOS & CKD \cite{ckd} & R50 & R101 & 1x
		& 39.49 & 58.09 & 42.77 & 23.40 & 43.26 & 50.89 \\
		FCOS & SKD \cite{skd} & R50 & R101 & 1x
		& 38.03 & 56.57 & 41.04 & 23.03 & 41.83 & 48.57 \\
		FCOS & Defeat \cite{defeat} & R50 & R101 & 1x
		& 39.25 & 57.97 & 42.25 & 23.52 & 43.09 & 49.82 \\
		FCOS & DSIG \cite{dsig} & R50 & R101 & 1x
		& 38.49 & 56.29 & 41.72 & 23.81 & 42.15 & 48.05 \\
		FCOS & Ours & R50 & R101 & 1x
		& \textbf{42.52} & 61.04 & 45.96 & 25.90 & 46.61 & 55.02 \\
		FCOS & $\dagger$GID \cite{gid} & R50 & R101 & 2x
		& 42.00 & 60.40 & 45.50 & 25.60 & 45.80 & 54.20 \\
		FCOS & AKD \cite{akd} & R50 & R101 & 2x
		& 42.30 & 61.00 & 45.50 & 26.80 & 45.90 & 54.50 \\
		FCOS & Ours & R50 & R101 & 2x
		& \textbf{43.06} & 61.89 & 46.49 & 27.65 & 47.15 & 55.71 \\
		\shline
		Faster R-CNN & Baseline & R50 & - & 1x
		& 38.03 & 58.91 & 41.13 & 22.21 & 41.46 & 49.22 \\
		Faster R-CNN & FGFI \cite{fgfi} & R50 & R101 & 1x
		& 38.85 & 59.62 & 42.16 & 22.68 & 42.20 & 50.48 \\
		Faster R-CNN & TAD \cite{tkd} & R50 & R 101 & 1x
		& 39.89 & 60.03 & 43.19 & 23.73 & 43.23 & 52.34 \\
		Faster R-CNN & CKD \cite{ckd} & R50 & R101 & 1x
		& 38.29 & 59.09 & 41.51 & 22.44 & 41.57 & 49.74 \\
		Faster R-CNN & SKD \cite{skd} & R50 & R101 & 1x
		& 38.29 & 58.47 & 41.83 & 21.95 & 41.67 & 49.33 \\
		Faster R-CNN & Defeat \cite{defeat} & R50 & R101 & 1x
		& 38.94 & 59.02 & 42.23 & 22.19 & 42.19 & 51.47 \\
		Faster R-CNN & DSIG \cite{dsig} & R50 & R101 & 1x
		& 40.57 & 61.15 & 44.38 & 24.17 & 44.06 & 52.80 \\
		Faster R-CNN & Ours & R50 & R101 & 1x
		& \textbf{40.64} & 61.38 & 44.31 & 24.03 & 44.16 & 53.03 \\
		Faster R-CNN & $\dagger$GID \cite{gid} & R50 & R101 & 2x
		& 40.20 & 60.70 & 43.80 & 22.70 & 44.00 & 53.20 \\
		Faster R-CNN & AKD \cite{akd} & R50 & R101 & 2x
		& 41.50 & 62.10 & 45.40 & 24.60 & 45.60 & 53.70 \\
		Faster R-CNN & Ours & R50 & R101 & 2x
		& \textbf{41.54} & 61.95 & 45.36 & 24.41 & 45.04 & 53.95 \\
	\end{tabular}
	\vspace{-2mm}
	\caption{Performance comparison with state-of-the-art methods in \textbf{Object Detection} \emph{box} AP on COCO \texttt{val} \wrt one-stage anchor-based RetinaNet \cite{retina}, one-stage anchor-free FCOS \cite{fcos}, and two-stage Faster R-CNN \cite{fasterrcnn}.}
	\label{table:sotacompare}
	\vspace{-1mm}
\end{table*}

We compare our method with state-of-the-arts in recent years, and show results in Table~\ref{table:sotacompare}. For fairness, we implement all methods based on the Detectron2 codebase except those marked with $\dagger$, where results are from the original papers. In one-stage anchor-based RetinaNet, our 1x method achieves superior results, outperforming all previous methods by 1.89 AP on average. Our 2x students also surpass previous state-of-the-arts \cite{akd} by 0.92 AP. Also, for one-stage anchor-free FCOS, our method achieves great gain over previous SOTAs of 3.56 point AP on average. And we improve the previous SOTA \cite{akd} by 0.76 AP on 2x results.

Since anchor-free methods rely more on rich information in each pixel to complete both classification and regression without box priors, they benefit more from our method that can better distill fine-grained pixel-level information.
On RetinaNet \cite{retina} and FCOS \cite{fcos}, our method outperforms the region-based ones \cite{fgfi, tkd, defeat} by 2.64 AP on average, indicating the high applicability of our method than solutions that nominate regions. Compared with relation-based methods \cite{dsig, gid}, ours yields 1.72 AP gain, manifesting that the topological distance is more effective than instance-relations in one-stage detectors. We also achieve 0.84 AP higher than backbone-based method \cite{akd}, and 3.04 AP higher than segmentation-based methods \cite{skd, ckd}, which manifest effectiveness of our design in the detector head.

Even for the two-stage Faster R-CNN, we still accomplish comparable results on students and slightly better results than state-of-the-arts \cite{akd, dsig}. It reveals the great potential of our method to generally improve detector distillation.

\subsection{Experiments for Instance Segmentation}
We also extend our method to the instance segmentation task and manifest our improvement on predicted masks. We adopt two-stage box-based Mask R-CNN \cite{maskrcnn} as well as one-stage box-free SOLOv2 \cite{solov2} as our network and evaluate the same student-teacher pairs as before. Models are trained on COCO2017 images that contain annotated masks. We report the standard evaluation metric AP$_{mask}$ based on \textit{Mask} IoU. All other training setting is the same as that described in Section~\ref{sec:training_detail}.  Results are shown in Table~\ref{table:instseg}. For Mask R-CNN, on four distillation pairs, we achieve 3.0 AP$_{box}$ and 2.3 AP$_{mask}$ gain on average over the baseline.
For SOLOv2, all students are improved about 2.33 AP$_{mask}$ on average, indicating that our method is also beneficial for both one- and two-stage mask prediction.

\subsection{Ablation Studies}

\subsubsection{Component Analysis}
\begin{table*}[t]
	\tablestyle{3.5pt}{1.02}
	\begin{minipage}[!t]{1.45\columnwidth}
		\begin{tabular}{l|cccc|x{22}x{22}x{22}|x{22}x{22}x{22}}
			& Baseline & $L_{anchor}$ & $L_{distance}$ & $L_{loc}$ & AP &  AP$_{50}$ & AP$_{75}$ & AP$_S$ &  AP$_M$ &  AP$_L$\\
			\shline
			FCOS & \checkmark &  &  & 
			& 39.34 & 58.09 & 42.64 & 24.16 & 43.37 & 49.77  \\
			FCOS & \checkmark & \checkmark &  & 
			& 41.65 & 60.27 & 45.04 & 25.03 & 45.51 & 53.97 \\
			FCOS & \checkmark & & \checkmark & 
			& 40.39 & 59.25 & 43.66 & 24.88 & 44.20 & 51.60 \\
			FCOS & \checkmark & \checkmark & \checkmark & 
			& 41.97 & 60.85 & 45.41 & 25.62 & 46.03 & 54.47 \\
			FCOS & \checkmark & \checkmark & \checkmark & \checkmark
			& \textbf{42.52} & 61.04 & 45.96 & 25.90 & 46.61 & 55.02 \\
		\end{tabular}
	\end{minipage}
	\begin{minipage}[!t]{0.5\columnwidth}
		\begin{tabular}{l|c|x{22}}
			& Loss on dense head     & AP \\
			\shline
			FCOS  & L$_{2}$   &  40.76  \\
			FCOS  & L$_{anchor}$ & \textbf{42.52}  \\
		\end{tabular}
		\vspace{0.5mm}
		\begin{tabular}{l|c|x{22}}
			& Loss on Bbox layer       & AP \\
			\shline
			FCOS  & L$_{2}$   &  42.06 \\
			FCOS  & L$_{loc}$ & \textbf{42.52}  \\
		\end{tabular}
	\end{minipage}
	\vspace{-3mm}
	\caption{Ablation study of loss component analysis in \textbf{Object Detection} \textit{box} AP on COCO \texttt{val} \wrt FCOS \cite{fcos} with ResNet50.}
	\label{table:ablation}
\end{table*}

As shown in Table~\ref{table:ablation}, we conduct experiments on multiple combinations of different losses in our method to manifest that each of them makes a difference. 

\vspace{-0.01in}
\paragraph{Anchor Loss:} 
Implementing $L_{anchor}$ brings about 2.31 AP to student baseline, whereas the direct pixel-pixel matching only achieves 1.42 AP. It means when the dense pixels are responsible for prediction tasks, it is necessary to summarize the categorical information among all pixels rather than mimicking global feature maps. With the distillation of $L_{anchor}$, information in large size feature maps avoids distillation in an imbalanced way since it concentrates on the category anchor.    

\vspace{-0.01in}
\paragraph{Distance Loss:}
Distance Loss $L_{distance}$ outperforms the baseline by about 1.05 AP, which means that in dense detectors, correlations among pixels are important to form the dense topological space. Instead of modeling dense relations like \cite{skd}, the $L_{distance}$ restricts the distance between pixel itself and each category anchor of student and teacher, which makes it better regularization for the student.

\vspace{-0.01in}
\paragraph{Location Distribution Alignment:}
Adding $L_{loc}$ improves FCOS-ResNet50 to 42.52 AP. It does help align the localization information of student and teacher, which reveals that the pixels in bounding box layer conforms to a kind of distribution.  
Also, applying L2 loss on the bounding box layer does no benefit to performance, which shows that matching the distribution in the bounding box layer is more effective than straightforward pixel-pixel distillation.

\subsubsection{Hyper-parameters Sensitivities}

\begin{figure}[t]
	\begin{center}
		\includegraphics[width=1.0\linewidth]
		{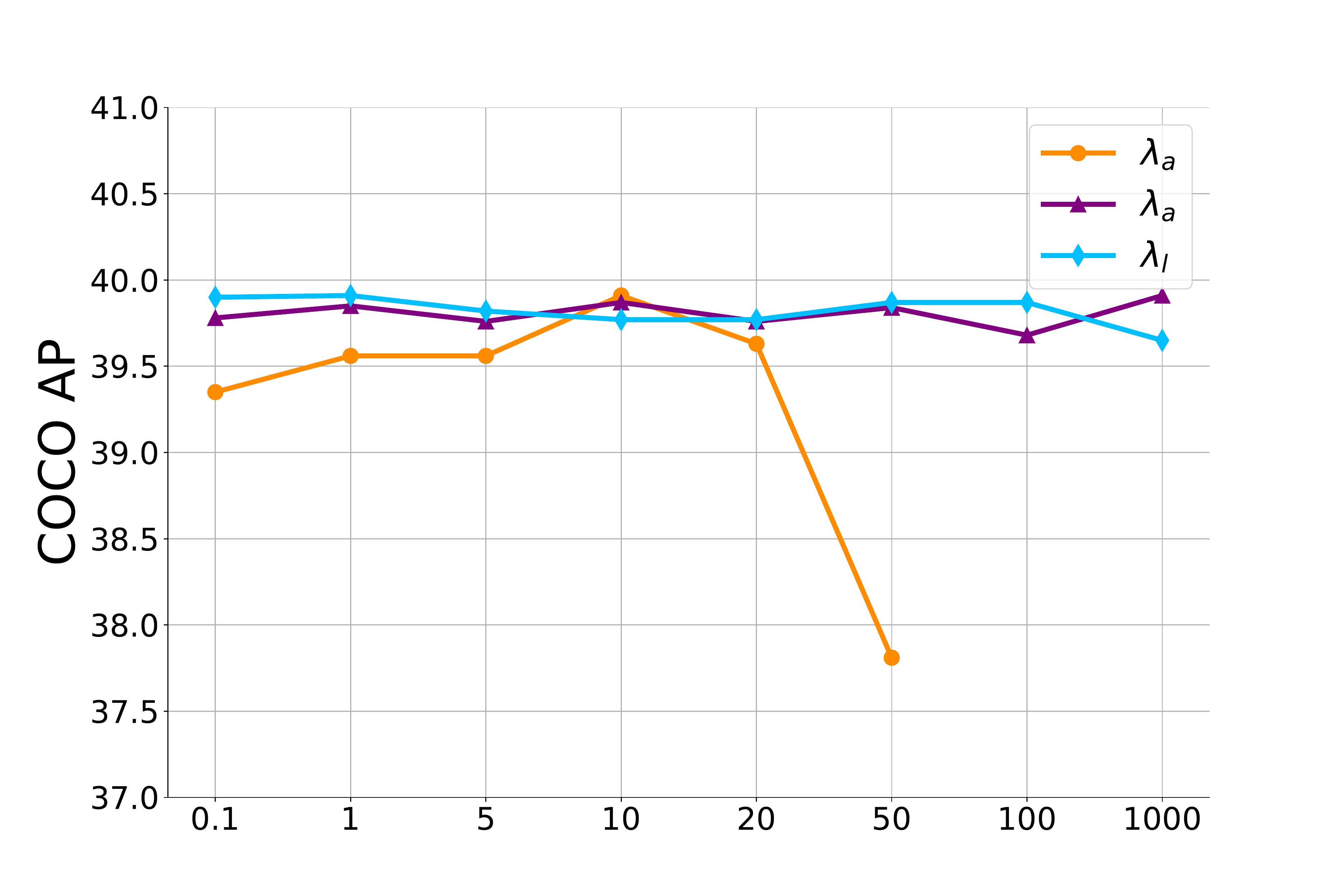}
	\end{center}
	\vspace{-8mm}
	\caption{Ablation study of loss penalty coefficients on COCO \texttt{val} \wrt RetinaNet \cite{retina} with ResNet50.}
	\vspace{-3mm}
	\label{fig:lw_sens}
\end{figure}

\paragraph{Loss penalty coefficients:}
We test sensitivities of three loss penalty coefficients $\lambda_a$, $\lambda_d$ and $\lambda_l$ in Eq.~(\ref{eq:overallloss}) (see Figure~\ref{fig:lw_sens}). It shows these coefficients are robust in a large range, which verifies the stability of our method.

\vspace{-0.01in}
\paragraph{Temperatures:}
We test the logit temperatures $\tau_d$ and $\tau_l$ of KLD losses in Eqs.~(\ref{eq:distanceloss})-(\ref{eq:locloss}), and show results in Table~\ref{table:ablation_temp}. The performance is robust within the range of 0.01 and 5.0. 

\begin{table}[t]
	\vspace{-3mm}
	\tablestyle{3.5pt}{1.05}
	\begin{tabular}{c|x{22}x{22}x{22}x{22}x{22}x{22}x{22}}
		Values & 0.01 & 0.1 & 0.2 & 0.5 & 1.0 & 2.0 & 5.0 \\
		\shline
		$\tau_d$ & 39.83 & 39.91 & 39.74 & 39.90 & 39.66 & 39.88 & 39.81 \\
		$\tau_l$ & 39.78 & 39.91 & 39.83 & 39.75 & 39.74 & 39.81 & 39.84 \\
	\end{tabular}
	\vspace{-2mm}
	\caption{Ablation study of logit temperatures in \textbf{Object Detection} \textit{box} AP on COCO \texttt{val} \wrt RetinaNet \cite{retina} with ResNet50.}
	\label{table:ablation_temp}
	\vspace{-3mm}
\end{table}

\section{Concluding Remarks}
\label{sec:discuss}

In this paper, we have proposed SEA (SEmantic-Aware Alignment) distillation for object detectors. To bridge the gap between one- and two-stage detectors distillation, we regard each pixel as instances and design Category Anchor to summarize the categorical information in scene image and deal with the drastic imbalance in dense pixels. Based on that, we model semantic relations and sparsify them to make distillation more structural and complete. Also, we effectively align the localization distribution in the under-studied bounding box branch between student and teacher. Extensive experiments manifest the effectiveness and robustness of our method regarding both object detection and instance segmentation distillation tasks.  

\paragraph{Limitations:}
The general limitation lies in the nature of distillation that the teacher is inevitably needed to transfer its knowledge to student. While the distillation is mainly for small student models, it is hard to find a proper teacher for large-model students.

\paragraph{Societal Impact:}
Our method compresses the huge teacher models to tiny student models while maintaining the accuracy, which saves computational cost and environmental resource consumption.
Our method has no explicit negative societal impact, except that in distillation procedure, we spend more resource training than the original student detector and we need to prepare the pre-trained teacher model.


{\small
\bibliographystyle{ieee_fullname}
\bibliography{cvpr}
}

\end{document}